\documentclass[sigconf]{acmart} 
\usepackage{hyperref} 
\usepackage{tikz}
\usepackage{mathrsfs}
\usepackage{amsmath,amsfonts}
\usepackage{bbm}
\usepackage{bbding}
\usepackage{multirow}

\usepackage{algorithmicx,algorithm}
\usepackage{algpseudocode} 
\usepackage{graphicx}
\usepackage{subfigure}

\usepackage{textcomp}
\usepackage{xcolor}

\usepackage{xspace}

\renewcommand\footnotetextcopyrightpermission[1]{}
\usepackage{tcolorbox}
\usepackage{soul}
\usepackage{pifont}
\begin{document}

\title{Training Data Attribution: Was Your Model Secretly Trained On Data Created By Mine?}

\author{Likun Zhang, Hao Wu, Lingcui Zhang, Fengyuan Xu, Jin Cao, Fenghua Li, Ben Niu} 
\authornote{Corresponding author.}
\begin{abstract}
 

The emergence of text-to-image models has recently sparked significant interest, but the attendant is a looming shadow of potential infringement by violating the user terms. 
Specifically, an adversary may exploit data created by a commercial model to train their own without proper authorization. 
To address such risk, it is crucial to investigate the attribution of a suspicious model's training data by determining whether its training data originates, wholly or partially, from a specific source model.
To trace the generated data, existing methods require applying extra watermarks during either the training or inference phases of the source model.
However, these methods are impractical for pre-trained models that have been released, especially when model owners lack security expertise.
To tackle this challenge, we propose an injection-free training data attribution method for text-to-image models.
It can identify whether a suspicious model's training data stems from a source model, without additional modifications on the source model.
The crux of our method lies in the inherent memorization characteristic of text-to-image models.
Our core insight is that the memorization of the training dataset is passed down through the data generated by the source model to the model trained on that data, making the source model and the infringing model exhibit consistent behaviors on specific samples.
Therefore, our approach involves developing algorithms to uncover these distinct samples and using them as inherent watermarks to verify if a suspicious model originates from the source model.
Our experiments demonstrate that our method achieves an accuracy of over 80\% in identifying the source of a suspicious model's training data, without interfering the original training or generation process of the source model.
\end{abstract}
\maketitle


\section{Introduction}

Text-to-image generation systems based on diffusion models have become popular tools for creating digital images and artistic creations~\cite{Ramesh2022dalle2, Rombach22sdm}. 
Given an input prompt in natural language, these generative systems can synthesize digital images of high aesthetic quality. 
Nevertheless, training these models is quite an intensive task, demanding substantial amounts of data and training resources.
They make such models valuable intellectual properties for model owners, even if the model structures are usually public.

\begin{figure}[t]
  \centering
  \includegraphics[width=0.45\textwidth]{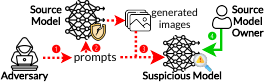}
  \caption{The task of training data attribution. An adversary may produce some prompts ({\color{red}\ding{182}}) and query the source model ({\color{red}\ding{183}}), then it collects generated images to train a model ({\color{red}\ding{184}}). The source model owner wants to investigate whether a model is trained on the data generated by the source model ({\color{green}\ding{185}}). Note that the suspicious model may be an innocent one. (See Section~\ref{sec_pre} for details.}
  \label{fig:scenario}
  \vspace{-10pt}
\end{figure}

One significant concern for such models is the unauthorized usage of their generated data~\cite{li23code}.
As illustrated in \autoref{fig:scenario}, an attacker could potentially query a commercial model and collect the data generated by the model, then use the generated data to train their personalized model. 
For simplicity in narration, we denote the attacker's model as the \textbf{suspicious model} and denote the commercial model as the \textbf{source model}.
This attack has already raised the alarm among commercial model developers. 
Some leading companies, e.g., MidJourney~\cite{MidJourney} and ImagenAI~\cite{imageai}, have explicitly stated in their user terms that such practices are not permitted, as shown in \autoref{fig:model_licenses}. 
It is crucial to investigate the relationship between the source model and the suspicious model. We term the task as training data attribution.

\begin{figure}[t]
    \centering
    \begin{tcolorbox}[title=\textbf{\small{The terms of services of MidJourney.}}]
    
    \small{... You may \hl{not use the Services for purposes of developing or offering competitive products or services}. You may not resell or redistribute the Services or access to the Service. ...}
    \end{tcolorbox}

    \begin{tcolorbox}[title=\textbf{\small{The terms of use policies of ImagenAI.}}]
    
    \small{... you may \hl{not copy, distribute, process, modify or create derivative works of any of the Service or any content}, either by yourself or by a third party on your behalf ...}
    \end{tcolorbox}
    
    \caption{User terms of commercial text-to-image models.\label{fig:model_licenses}} 
    \vspace{-10pt}
\end{figure}

To tackle the task, one may think of using the watermarking techniques to accomplish the task. 
Existing watermarking methods can generally be categorized into two types: one involves embedding watermarks in the training data during the model training phase \cite{Luo2023StealMA, zhao2023recipe, liu23watermarking}, and the other adds watermarks to the model outputs after training~\cite{li23code}, such that the generated data contains traceable watermark characteristics. 
However, there are two issues existing works do not fully address.
\textit{Firstly}, regarding feasibility, it remains unexplored whether a source model, once watermarked, can successfully pass its watermark to a suspicious model through generated data. 
\textit{Secondly}, regarding usability, watermark techniques could affect the generation quality of the source model \cite{zhao2023recipe}, either during or after training. Moreover, applying these techniques requires security knowledge, thereby raising the bar for practical application.

In this paper, our goal is to uncover the indicators naturally embedded within a source model, which can be transferred to any model trained on data produced by the source model. 
These inherent watermarks can reveal the relationship between the source and suspicious models. Unlike artificially injected watermarks, these inherent indicators do not require modifications to the model's training algorithm or outputs. This means that applying our attribution method approach won't compromise the model's generation quality and doesn't necessitate any security knowledge.

The rationale of our approach stems from the memorization phenomenon exhibited by text-to-image generation models. The memorization signifies a model's ability to remember and reproduce images of certain training samples when the model is prompted by the corresponding texts during inference~\cite{nips23UnderstandingCopying}. 
Research has shown that such memorization in generative models is not occasional.
On the contrary, models with superior performance and stronger generalization abilities demonstrate more notable memorization~\cite{nips23UnderstandingCopying}.

Though promising, applying the memorization phenomenon to achieve our goal is not straightforward.
Even if we manage to conduct a successful training data extraction on the suspicious model as proposed in \cite{Carlini23extracting}, the information we procure is the data generated by the source model.
Given that the generation space of the source model is vast, it becomes challenging to verify whether the extracted data was generated by the source model. 
We will detail the challenges in a formal manner in Section~\ref{sec_pre}.

In this paper, we propose a practical injection-free method to ascertain whether a suspicious model has been trained using data generated by a certain source model. Our approach considers both the instance-level and statistical behavior characteristics of the source model, which are treated as part of the inherent indicators to trace its generated data against unauthorized usage.
In particular, \textit{at the instance level}, we devise two strategies to select a set of key samples (in the form of text and image pairs) within the source model's training data. This set is chosen to maximize the memorization phenomenon. We then use the texts in these samples to query the suspicious model at minimal cost, evaluating the relationship between the two models based on the similarity of their outputs.
\textit{At the statistical level}, we develop a technique involving the training of several shadow models on the datasets that contain or do not contain the data generated by the source model. Then we estimate the metric distributions for data attribution with a high-confidence. 

Experimental results demonstrate that our instance-level attribution solution is reliable in identifying an infringing model with high confidence over $0.8$.
Even when the infringing model uses only a small proportion as 30\% of generated data, the attribution confidence is over 0.6, on par with the existing watermark-based attribution method.
The statistical-level solution achieves an overall accuracy of over 85\% in distinguishing the source of a suspicious model's training data.
 
Our main contributions are summarized as:
\begin{itemize}
    \item Focusing on the issue of user term violation caused by the abuse of generated data from pre-trained text-to-image models, we formulate the problem as training data attribution in a realistic scenario.
     To the best of our knowledge, we are the first to work on investigating the relationship between a suspicious model and the source model.
    
    \item We propose two novel injection-free solutions to attribute training data of a suspicious model to the source model at both the instance level and statistical level. 
    These methods can effectively and reliably identify whether a suspicious model has been trained on data produced by a source model.
    
   \item We carry out an extensive evaluation of our attribution approach. Results demonstrate its performance is on par with existing watermark-based attribution approach where watermarks are injected before a model is deployed.
\end{itemize}

The rest of the paper is organized as follows.
We introduce the background knowledge and related works in Section~\ref{sec_relat}.
Section~\ref{sec_pre} describes the preliminary and our assumptions.
Then, Section~\ref{sec_alg} presents our research question and the attribution approach in detail. 
Experimental evaluation results are reported in Section~\ref{sec_eva}. 
Finally, we conclude this paper in Section~\ref{sec_concl}.





\section{Background \& Related Work}
\label{sec_relat}

\subsection{Text-to-Image Diffusion Model}
\label{sec:tti}
  
In general, a text-to-image model is a type of conditional generative model that aims to create images based on textual descriptions through generative models.
They are trained with data in the form of image-text pairs. 
In this paper, we take the currently state-of-the-art text-to-image model, i.e., Stable Diffusion (SD)~\cite{Rombach22sdm}, to prototype our method.
However, note that our approach can be applied to protecting other types of models. 
Stable Diffusion (SD)~\cite{Rombach22sdm} is a typical latent diffusion model (LDM). SD mainly contains three modules: (1) Text encoder module $W$: it takes a text prompt $P$, and encodes it into the corresponding text embedding $c = W(P)$; (2) Autoencoder module including an image encoder \scalebox{1.3}{$\varepsilon$} and decoder $D$: \scalebox{1.3}{$\varepsilon$} transforms an image $x$ into a latent representation space where $z=\scalebox{1.3}{$\varepsilon$}(x)$, while $D$ maps such latent representations back to images, such that $D($\scalebox{1.3}{$\varepsilon$}$(x))\approx x$; (3) Conditional diffusion module $\epsilon_\theta$ parameterized by $\theta$: a U-Net model~\cite{ronneberger15unet} trained as a noise predictor to gradually denoise data from randomly sampled Gaussian noise conditioned on the input text embedding.

The objective for learning such a conditional diffusion model (based on image-condition training pairs $(x,c)$) is as follows:
 \begin{equation}
 \label{eq_ldm}
     \mathcal{L}_{LDM}=\mathbb{E}_{\scalebox{1.1}{$\varepsilon$}(x), c, \epsilon\sim\mathcal{N}(0,1), t}\left[{||\epsilon-\epsilon_\theta(z_t, t, c)||^2_2}\right].
 \end{equation}
After the denoising, the latent representation $z$ is decoded into an image by $D$.

\subsection{Watermarking Techniques}

Recent studies suggest the use of watermarking techniques as a defense against misuse of generated data. 
These techniques help identify copy-paste models~\cite{zhao2023recipe, liu23watermarking} or models subjected to extraction attacks~\cite{jia21entangled, lv24mea}. 
Typically, these watermarks are embedded either in the model during the training phase or in the output during the generation stage.

\noindent \textbf{Watermarking during the training phase.}
One common approach involves using backdoor triggers as watermarks. 
This helps identify models that directly reuse source model weights~\cite{usenix18backdoor}. Recent studies have also shown that text-to-image diffusion models can be vulnerable to backdoor attacks~\cite{chen23Trojan, chou23backdoor, Struppek22text_backdoor, zhao2023recipe, liu23watermarking}. 
However, these trigger-based watermarks may be easily removed under model extraction attacks due to weight sparsity and the stealthiness of the backdoor. 
To combat this, Jia \textit{et al.}~\cite{jia21entangled} suggested intertwining representations extracted from training data with watermarks. 
Lv \textit{et al.}~\cite{lv24mea} advanced this idea for self-supervised learning models, loosening the requirement for victim and extracted models to share the same architecture.

\noindent \textbf{Watermarking during generation phase.}
It involves modifying the model outputs to embed the unique watermarks of the model owner.
For LLM-based code generation models, Li \textit{et al.}~\cite{li23code} designed special watermarks by replacing tokens in the generated code with synonymous alternatives from the programming language.
Consequently, any model resulting from an extraction attack will adopt the same coding style and produce watermarked code traceable to the original data source.

Currently, watermarking techniques have not yet been explored for their potential to tackle the training data attribution task (See Section~\ref{subsec_re_prob}).
Additionally, applying these techniques can lead to a drop in the quality of data generated by the model~\cite{zhao2023recipe}. 
Moreover, these techniques could reduce the quality of the data generated by the model~\cite{zhao2023recipe}, and they often require specialized security knowledge for implementation during model development. Our approach seeks to address these issues without compromising the quality of generated data or requiring developers to have a background in security.

\section{Preliminary}
\label{sec_pre}

\subsection{Problem Statement}
\label{sec_prob_state}

\underline{\textit{Source model.}} We denote the well-trained text-to-image source model as $\mathcal{M}_S$.
The source model is trained with a large amount of high-quality ``text-image'' pairs, denoted as $\{\mathtt{TXT}_t, \mathtt{IMG}_t\}$.
During the inference phase, it can generate an $\mathtt{img}$, given a text prompt $\mathtt{txt}$, i.e., 
\begin{equation}
    \mathtt{img} = \mathcal{M}_S(\mathtt{txt}) .
\end{equation}

 \underline{\textit{Aggressive infringing model.}} An aggressive adversary might aim to train its text-to-image model to offer online services for economic gain.
The adversary can easily obtain an open-source model architecture, which may be the same as the source model or may be not. 
The adversary does not have enough high-quality ``text-image'' pairs to train a satisfactory model. 
It can prepare the training dataset in the following manner. 
The adversary prepares a set of text $\mathtt{TXT}_A$, and it queries the $\mathcal{M}_S$ with the set of text, and collects the corresponding $\mathtt{IMG}_A$ generated by $\mathcal{M}_S$.
Then, the adversary trains its model $\mathcal{M}_A$ with the generated data pairs.
As the user terms reported in Figure~\ref{fig:model_licenses}, \textit{the adversary abuses the generated data, and the right of the source model is violated. }

\underline{\textit{Inconspicuous infringing model.}} There may be an adversary, who already has some training data, which is insufficient to train a satisfactory model. 
Hence, it also collected some generated data from the model $\mathcal{M}_S$, using the aforementioned method. 
Then, it mixes the generated infringing data with its own data for model training. 
We denoted the mixed dataset as $\{\mathtt{TXT}_M, \mathtt{IMG}_M\}^\rho$, where $\rho\in(0,1)$ refers to the proportion of samples that are generated by $\mathcal{M}_S$ in the mixed dataset, i.e., 
\begin{equation}
\rho = \frac{|\mathtt{IMG}_A |}{|\mathtt{IMG}_M|} .
\end{equation}

Note that when $\rho$ equals 1, the inconspicuous adversary becomes the aggressive adversary. Therefore, for simplicity, we use the following notations to represent these two types of adversaries, i.e.,
\begin{equation}
\mathcal{M}_A^\rho = 
\begin{cases} 
\text{Aggressive infringing model,} & \text{if } \rho = 1;\\
\text{Inconspicuous infringing model,} & \text{if } \rho \in (0, 1).
\end{cases}
\end{equation}

\underline{\textit{Innocent model.}} For the sake of rigorous narration, we define an innocent model, denoted as $\mathcal{M}_{In}$, which provides similar services as the source model, but its training data has no connection whatsoever with the data generated by the $\mathcal{M}_S$.

\underline{\textit{Suspicious model.}}
From the perspective of the owner of $\mathcal{M}_S$, facing any text-to-image model with similar functionality, they wouldn't know if this model has been trained using data generated by the source model. 
Thus, the owner would refer to this model as a suspicious model. Note that, this suspicious model could be an infringing model ($\mathcal{M}_A^\rho$) or an innocent model ($\mathcal{M}_{In}$).

\noindent \textbf{Our goals.} Our attribution method aims to achieve two goals: 
\begin{enumerate}
    \item Determine whether a suspicious model is an innocent model ($\mathcal{M}_{In}$) or an infringing model ($\mathcal{M}_A^\rho$). 
    \item Improve the robustness of our attribution method towards $\rho$. (Make the proposed method remain effective even when $\rho$ is small).
\end{enumerate}

\subsection{Assumptions}
\label{assumption}
Here we make some reasonable assumptions to better illustrate our working scenario. 

\noindent \textbf{About the source model and its owner.} The model architecture and training algorithm of model $\mathcal{M}_S$ can be open-source. 
The owner of the source model $\mathcal{M}_S$ does not have any security knowledge, so it neither watermarks any training data during the model training nor modifies the model output in the inference phase for watermarking purposes. 
The question of utmost concern for the model owner, as shown in Figure~\ref{fig:model_licenses} is whether the data generated by $\mathcal{M}_S$ has been used to train another model.
The source model owner has full knowledge of the model architecture and parameters and can access all training data of the $\mathcal{M}_S$. 

We hypothesize that the training process of the source model might involve both public-accessible data and private data. Consequently, the generated data may contain examples related to both public and private data. \textit{This paper discusses the attribution of generated data relevant to private data.}

\noindent \textbf{About the suspicious model.}
The suspicious model $\mathcal{M}$ is in a black-box setting. 
The suspicious model may share the same model architecture as the source model.
The functionality of the suspicious model is also provided, which is necessary for an ordinary user to use the suspicious model.
It only offers a query-only interface for users to perform the investigation.

\section{Methodology}
\label{sec_alg}

\subsection{Research Problem}
\label{subsec_re_prob}
\begin{figure}[t]
  \centering
  \includegraphics[width=0.4\textwidth]{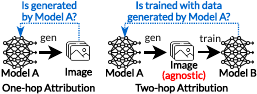}
  \caption{Our research question. The one-hop attribution is well-studied in the field of data watermarking. Our paper attempts to solve the two-hop attribution in real world generation setting.}
  \label{fig:research_question}
\end{figure}

We define the task of "determining whether a piece of data is generated by a particular model" as a \textit{one-hop data attribution}. This idea is illustrated in Figure~\ref{fig:research_question}.
The one-hop data attribution is gaining attention in both academia~\cite{liu23watermarking, zhao2023recipe} and industry circles~\cite{Rombach22sdm,Ramesh2022dalle2}. 
Checking the presence of a certain watermark on the generated data is a common one-hop data attribution procedure.

Our work focuses on \textit{two-hop attribution}, that is, we aim to determine if Model B has been trained using data generated by Model A. In this setting, the data generated by Model A cannot be enumerated, and the generated data is not embedded with watermarks.
This task has caught recent attention, and Han \textit{et al.}~\cite{han24model_attribution} made an initial exploration on whether the training data of a classification model is generated by a specific GAN model in the aggressive infringing setting as defined in \autoref{sec_pre}.

Compared to the existing effort, our work addresses a more challenging task under a real-world generation scenario.
\textit{First, we investigate a more realistic threat model.} 
We consider not only the aggressive infringing model but also an inconspicuous setting. We argue that the inconspicuous setting is more prevalent, especially when many developers can only collect a small amount of data to fine-tune their models instead of training from scratch.
\textit{Second, we examine more complex subjects.} Previous studies explored source models with simple GAN networks, and the suspicious model was a closed-vocabulary classification model. However, in our study, both the source model and the suspicious model are unexplored text-to-image diffusion models capable of managing open-vocabulary generation tasks, which makes them tougher to analyze.

\subsection{Design Overview}

As illustrated in Figure~\ref{fig:research_question}, within the two-hop attribution context, the generated data used to train Model B is agnostic. 
Therefore, to solve the two-hop data attribution, we must establish a connection between Model B and Model A. 
This is similar to works in the field of model extraction attacks~\cite{wu2022steal_vlm, ShaCVPR23steal_encoder, Luo2023StealMA}. 

Model extraction research demonstrates that it's possible to train a clone model (denoted as $\mathcal{M}_\texttt{clone}$) with an identical performance level to the target model (denoted as $\mathcal{M}_\texttt{target}$) using data generated by the latter. This can be mathematically represented as:
\begin{equation}
    | \mathcal{M}_\texttt{clone}(x) - \mathcal{M}_\texttt{target}(x) | < \epsilon ,
    \label{eq:extraction}
\end{equation}
where $x \sim \mathcal{X}$ is any input from the distribution $\mathcal{X}$, and $\epsilon$ is a small positive number, signifying the extraction error.

\begin{figure}[t]
  \centering
  \includegraphics[width=0.42\textwidth]{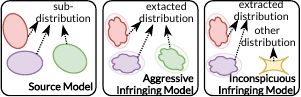}
  \caption{Our core insight. In the open-vocabulary generation task, the source model can generate data in different distributions. In the view of a model extraction attack, the infringing model may extract all or part of the distributions of the source model. The $\epsilon$ in \autoref{eq:extraction} indicates the difference between the extracted distribution and the source distribution.}
  \label{fig:core_insight}
  \vspace{-15pt}
\end{figure}

Inspired by the model extraction tasks, we describe the two-hop attribution task in Figure~\ref{fig:core_insight}.
An infringing model might either completely (i.e., aggressive setting) or partially (i.e., inconspicuous setting) duplicate the source model's distribution. 
Our primary insight in addressing this concern is to \textbf{identify the extracted distribution} present in the suspicious model. To achieve this, we assess the relationship between the behaviors of the source and suspicious models, both at instance and statistical levels.

\noindent \textbf{At instance level}, we aim to identify an infringing model through measuring the attribution confidence on a set of instances. Guided by Equation~\ref{eq:extraction}, we use a set of key samples to query both the source and suspicious models, subsequently measuring the similarity of their responses. 
The challenge lies in the selection of key samples. We will elaborate on this in Section~\ref{subsec_instance_level}.

\noindent \textbf{At statistical level}, we aim to measure the behavior differences between the innocent model and infringing model. We hypothesize that, given inputs from the source model's distribution, there will be a significant performance gap between the infringing and innocent models. The challenge here is to develop a technique that accurately measures this difference. We will delve into this in Section~\ref{sec:shadow}.

The performance of the instance level solution relies on the ability to find samples that can accurately depict the distribution of of the source models' training data. It has superior interpretability. 
While the statistical level solution falls short in interpretability, it enables a more comprehensive attribution, and hence a superior accuracy. 
Therefore, in practice, we recommend users choose according to their specific requirements.

\subsection{Instance-level Solution}
\label{subsec_instance_level}

The core of the instance-level solution is to capture the shared sub-distributions between the source and suspicious models (Refer to \autoref{fig:core_insight}).
In this context, we use $\{ \mathcal{X}_1, \ldots, \mathcal{X}_n \}$ to denote sub-distributions of the source model. 
The suspicious model's sub-distributions, which are shared with the source model, are represented as $\{ \mathcal{X}_1, \ldots, \mathcal{X}_m \}$.
It's important to note that when $m$ equals $n$, the suspicious model is considered an aggressive infringing model. If $m$ is less than $n$, it signifies an inconspicuous infringing model. Conversely, if $m$ equals 0, implying the suspicious model shares no sub-distribution with the source model, it is deemed an innocent model.
As assumed in Section~\ref{assumption}, the training data of the source model is private to the model owner, meaning others cannot access these data or any data from the same distribution through legitimate means.

The instance level solution can be formalized follows:
\begin{equation}
    \texttt{conf} = f(\mathcal{M}_S(x), \mathcal{M}(x)), ~\forall x \sim \mathcal{X}_i, ~ i \in \{1, \ldots, m\},
\label{eq:conf_f}
\end{equation}
where $\texttt{conf}$ is the confidence of whether the suspicious model $\mathcal{M}$ is an infringing one.
The formulation indicates two problems:
\textit{1)} how to \textit{prepare the input} $x$, since sampling from the distribution $\mathcal{X}_i$ cannot be exhaustive.
\textit{2)} how to \textit{design the attribution metric} $f$.
Next, we introduce two strategies to prepare the attribution input, and the detailed design of the attribution metric. 

\begin{figure}[!t]
  \centering
  \includegraphics[width=0.42\textwidth]{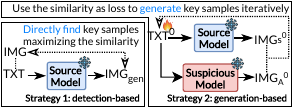}
  \caption{Two strategies for key samples preparation in instance-level solutions. Strategy 1 is detection-based, which aims to directly select key samples from the source model's training dataset. Strategy 2 is generation-based, which aims to synthesize key samples by maximizing the similarity between source and suspicious models. Note that in both strategies, no model update is needed.}
  \label{fig:two_strategies}
  \vspace{-15pt}
\end{figure}

\subsubsection{Attribution Input Preparation}

The idea behind preparing input data is if a set of instances $X$ can minimize the generation error of the source model $\mathcal{M}_S$, then these instances $X$ are most likely to belong to a sub-distribution learned by $\mathcal{M}_S$. 
Consequently, if these instances $X$ also minimize the generation error on a suspicious model, it suggests that this model has also been trained on the same sub-distribution. 
This leads to a conclusion that the suspicious model infringe on the source model, as we assume that only the source model owner holds data in this sub-distribution.
This assumption is reasonable and practical.
If an instance is easily obtained from a public distribution and not private to $\mathcal{M}_S$'s owner, there is no strong motivation to trace the usage.
Since our instance-level approach provides good interpretability, we can manually select those private instances from the instances prepared by our method for further investigation.
We term these instances as \textit{key samples}.

We develop two strategies to prepare \textit{key samples}, namely, a \textit{detection-based strategy} and a \textit{generation-based strategy}. 
We illustrate these two strategies in \autoref{fig:two_strategies}.
The detection-based strategy aims to identify a core set within the training dataset of $\mathcal{M}_S$ that minimizes generation error, which serves as representative samples of the model's distribution.  
This strategy is quick and does not require any training. 
The generation-based strategy focuses on creating samples from the source model $\mathcal{M}_S$ that can minimize the generation error. 
that can minimize the generation error. This strategy offers a broader sample space and superior accuracy compared to the detection-based strategy. 
Let's detail how these strategies work.

\noindent \textbf{Detection-based strategy.}
In this strategy, we start by feeding all text prompts $\mathtt{TXT}$ from the source model's training dataset into the source model $\mathcal{M}_S$.
From this, we generate images $\mathtt{IMG}_{gen}$. 
Next, we use the SSCD score~\cite{Pizzi_2022_sscd} to compare the similarity between $\mathtt{IMG}_{gen}$ and their ground-truth images $\mathtt{IMG}_{gt}$.
The SSCD score is the state-of-the-art image similarity measurement widely used in image copy detection\cite{cvpr23artForgery, nips23UnderstandingCopying}.  
We select $N$ instances with the largest similarity scores as key samples:
\begin{equation}
\mathtt{TXT}_k = \text{Top}_N \left\{ \text{SSCD}(\mathcal{M}_S(txt_i), img_i)|1\leq i\leq |\mathtt{TXT}| \right\},
\end{equation}
where $txt_i \in \mathtt{TXT}$, $img_i \in \mathtt{IMG}_{gt}$.
The key samples in $\mathtt{TXT}_k$ are taken as the attribution input.

\noindent \textbf{Generation-based strategy.}
In a text-to-image model, there are two components: the text encoder and the image decoder. For this particular strategy, we begin by randomly selecting a group of text prompts from the source model's training dataset. We refer to these as seed prompts.
Each selected text input (which we denote as $\texttt{txt}$) is comprised of $n$ tokens, i.e., $\texttt{txt} = [tok_1, tok_2, \ldots, tok_n]$.
The next step is to use the source model's text encoder to convert each token of $\texttt{txt}$ into an embedded form, producing $\texttt{c} = [c_1, c_2, ..., c_n] $.
After this embedding phase, we optimize $\texttt{c}$ over $T$ iterations to obtain an updated embedding, $\texttt{c}'$,
The optimization's objective is minimizing the reconstruction loss given by \autoref{eq_ldm} between the generated and ground-truth image.

Upon reaching convergence, we transform the optimized continuous text embedding $\texttt{c}'$ back to discrete token embeddings.
To do this, we find the nearest word embedding (referred to as $\texttt{c}^*$ in the vocabulary. 
However, because we carry out optimization at the word level, some of the resulting optimized embeddings may not make sense. To counteract this issue, we apply post-processing to the identified embeddings.
We calculate the hamming distance between the located embedding $\texttt{c}^*$ and its matching seed embedding $\texttt{c}$. 
We then retain the top-$N$ found embeddings, those with the smallest hamming distances.
Lastly, using the one-to-one mapping between the word embedding and the token in our vocabulary, we generate the attribution input $\texttt{txt}^*$.


\subsubsection{Attribution Metric for Instance Level Solution}
\label{metric_pr}
Now we use the similarity between the output of the source and suspicious model conditioned by the key samples to instantiate the metric $f$ in \autoref{eq:conf_f}.

Specifically, given the suspicious model, we query it with each text prompt $\mathtt{txt}_k$ from the key samples to obtain the corresponding generated image $\mathtt{img}_k$.
Similarly, we query the possible source model to obtain the corresponding generated image $\mathtt{img}'_k$.
We calculate the output distance as follows:
\begin{equation}
    d_k=||\mathtt{img}_k-\mathtt{img}'_k||_2
\end{equation}

Based on the distance between each image pair, we define the attribution metric $f$ in Equation~\ref{eq:conf_f}:
\begin{equation}
\label{eq:conf}
f \coloneq \texttt{conf}=\frac{1}{N}\sum_{j=1}^k\mathbbm{1}\left(d_k<\delta_0\right),
\end{equation} 
where $\mathbbm{1}(\cdot)$ is the indicator function.
It measures the confidence of whether the suspicious model infringes the source model.
The higher the $\texttt{conf}$ value, the more possible for a suspicious model to be an infringing model.
If the value of $\texttt{conf}$ surpasses a certain threshold $\delta$, we tend to assert that the suspicious model is indeed infringing the source model.
The $\delta_0$ is a hyper-parameter. We conduct analysis in \autoref{ablation_study} to ascertain its optimal value. 
In our context, we suggest setting $\delta_0=0.15$, independent of the type of source models.

\subsection{Statistical-level Solution}
\label{sec:shadow}

At the statistical level, we aim to capture the behavior differences among the source model ($\mathcal{M}_S$), the infringing model ($\mathcal{M}_A^\rho$), and the innocent model ($\mathcal{M}_{In}$). 
However, quantifying these differences directly poses a substantial challenge. Therefore, we opt for a parameterized approach.
Suppose a set of text prompts, denoted as $\mathtt{TXT}^p$. 
This set of prompts is used to query each model, which in turn generates $\mathtt{IMG}_S^p$, $\mathtt{IMG}_A^p$, and $\mathtt{IMG}_{In}^p$, accordingly.
Given the deterministic nature of the inference process, any differences observed among the resulting data can be extrapolated to imply distinctions among the models themselves.

Here, we formalize the statistical level solution as follows:
\begin{equation}
    \texttt{res} = f_D(\mathcal{M}(\mathtt{TXT}^p)),
\end{equation}
where $f_D$ is a discriminator model, e.g., a CNN model for the image classification task. 
If $\texttt{res} == 1$, then $\mathcal{M}$ is designated as infringing; if $\texttt{res} == 0$, it's classified as innocent. 
The $\mathtt{TXT}^p$ is prepared using the key samples selection method presented in Section~\ref{subsec_instance_level}.
Next, we describe how to train the $f_D$ in a supervised approach. 

We leverage the shadow model technique from the membership inference attack~\cite{shokri2017membership} to gather the labeled training data for $f_D$. 
It involves the following steps:
\begin{enumerate}
\item \textit{Data preparation for the shadow model.}
We select a set of text prompts, denoted as $\texttt{TXT}$, from the source model's training dataset, and collect corresponding images, represented as $\texttt{IMG}_{gt}$.
The $\texttt{TXT}$ is then fed into the source model and an innocent model to generate $\texttt{IMG}_{s}$ and $\texttt{IMG}_{in}$ respectively.
    
\item \textit{Innocent shadow model training.}
From datasets $\{\texttt{TXT}, \texttt{IMG}_{gt}\}$ and $\{\texttt{TXT}, \texttt{IMG}_{in}\}$, we randomly select $n$ samples, without replacement, iteratively repeat $s$ times to form $s$ subsets. 
We subsequently train $s$ innocent shadow models with the selected subsets.
It's worth noting that the use of samples from the dataset $\{\texttt{TXT}, \texttt{IMG}_{in}\}$ serves to minimize the effect of generated data.

\item \textit{Infringing shadow model training.}
From  $\{\texttt{TXT}, \texttt{IMG}_{gt}\}$ and $\{\texttt{TXT}, \texttt{IMG}_{s}\}$, we select samples based on the generation ratio defined by $\rho$. 
We randomly select $\rho \times n$ samples from the former dataset and select $(1-\rho) \times n$ samples from the latter. 
Here $\rho$ is randomly chosen within $(0,1]$.
We repeat the above sampling for $s$ times to obtain $s$ subsets for training $s$ infringing shadow models.
The purpose of sampling with different $\rho$ values is to enhance our method's robustness against inconspicuous adversaries.
    
\item \textit{$f_D$'s training dataset construction.}
We query each shadow model with $N$ texts in key samples and collect the corresponding images.
We label images generated by innocent shadow models as $0$, and label the outputs of infringing shadow models as $1$. 
These outputs and labels form a binary attribution dataset, which is partitioned into a training split and a testing split.
\end{enumerate}
Finally, we train the discriminator $f_D$ on the training split and assess its performance on the testing split.

\section{Experimental Evaluation}
\label{sec_eva}


In this section, we will first outline our experimental procedures. Then, we demonstrate if the proposed method can attain the objectives identified in Section~\ref{sec_prob_state}. Finally, we complete an ablation study and discuss strategies for selecting optimal hyperparameters.

\subsection{Settings}

\noindent\textbf{Text-to-image models.}
We use Stable Diffusion~\cite{Rombach22sdm} with the Stable-Diffusion-v1-5 (SD-v1)~\cite{sd15} and Stable-Diffusion-v2-1 (SD-v2)~\cite{sd21} checkpoints as the pre-trained models.


\noindent \textbf{Datasets.}
We select two widely adopted caption-image datasets.

\underline{\textit{1) CelebA-Dialog-HQ (CelebA)~\cite{jiang21celebaDialog}}}: a large-scale visual-language face dataset with 30,000 high-resolution face images with the size of $1024 \times 1024$ selected from the CelebA dataset.
Accompanied by each image, there is a caption that describes five fine-grained attributes including Bangs, Eyeglasses, Beard, Smiling, and Age.

\underline{\textit{2) Google's Conceptual Captions (CC3M)~\cite{CCdataset18}}}: a new dataset consisting of 3.3M images annotated with captions. We use its validation split which consists of 15,840 image/caption pairs. 
In contrast with the curated style of other image caption annotations, Conceptual Caption images and their descriptions are harvested from the web, and therefore represent a wider variety of styles. 

\noindent \textbf{Source model construction.}
We construct the source models by directly using pre-trained or consequently finetuning them on the above datasets. 
For the training data for finetuning, we randomly select 3000 samples from each dataset and resize them into $512\times 512$. 
We finetune each pre-trained model on each dataset for a total of 3000 iterations with a constant learning rate of 2e-6 and batch size of 2. 
We denote these source models as: SD-v1, SD-v2, SD-v1-CelebA, SD-v2-CelebA, SD-v1-CC3M, SD-v2-CC3M.

\noindent \textbf{Suspicious model construction.}
While pre-training and fine-tuning both raise concerns about IP infringement, fine-tuning has a more severe impact. 
Compared to pre-training, fine-tuning is highly convenient and efficient, allowing for many unauthorized uses without much resource restriction.
Thus we built each infringing model by finetuning a pre-trained model on $500$ training samples, where a proportion of $\rho$ of them are generated by a source model, while the rest are sampled from the real data. 
We follow the above pipeline to build the innocent models by setting $\rho=0$.


\noindent \textbf{Baselines.}
Note that our work is the first to address the problem in training data attribution in the text-to-image scenario, and therefore, there is no directly related work. Hence, we have designed two methods to comprehensively demonstrate our effectiveness.

 \underline{\textit{Baseline 1: Watermark-based data attribution}}. This baseline injects watermarks into the training data.
More specifically, as proposed in \cite{Luo2023StealMA}, by encoding a unique 32-bit array into the images generated by the source models, the infringing models trained on such watermarked data will also generate images in which the watermark can be detected. 
We believe watermark-injection based method showcasing the best attribution ability. 

\underline{\textit{Baseline 2:  Random selection-based data attribution}}.
This baseline adopts the similar idea with of our instance-level solution, but does not use the Strategy 1 and Strategy 2 we proposed for data attribution. 
Specifically, we randomly select $N$ training samples from the source model's training dataset as the attribution input. 
This serves as a baseline to demonstrate a straightforward attribution.


\noindent \textbf{Evaluation Metrics.}
We use the Accuracy, Area Under Curve (AUC) score, and TPR@10\%FPR~\cite{carlini22firstprinciple} to evaluate the accuracy and reliability of the attribution methods.
TPR@10\%FPR measures the true-positive rate (TPR) at a low false-positive rate (FPR).

\begin{figure}[!]
    \centering
    \includegraphics[width = 0.34\textwidth]{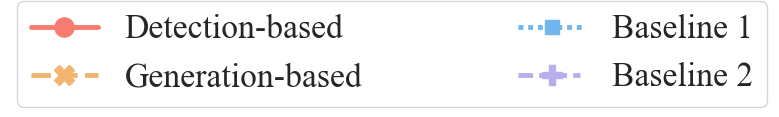}\vspace{-2.8mm}
    \subfigure[SD-v1]{
    \centering\includegraphics[width = 0.22\textwidth]{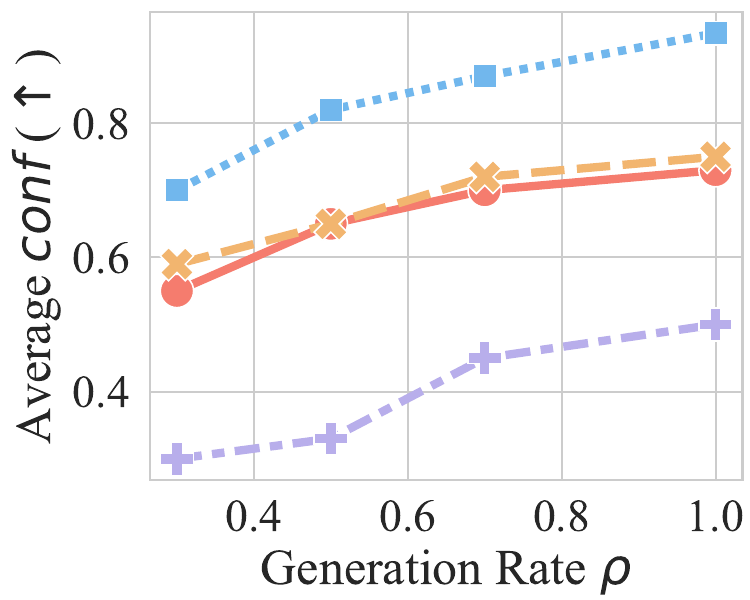}}
    \quad
    \subfigure[SD-v2]{\includegraphics[width = 0.22\textwidth]{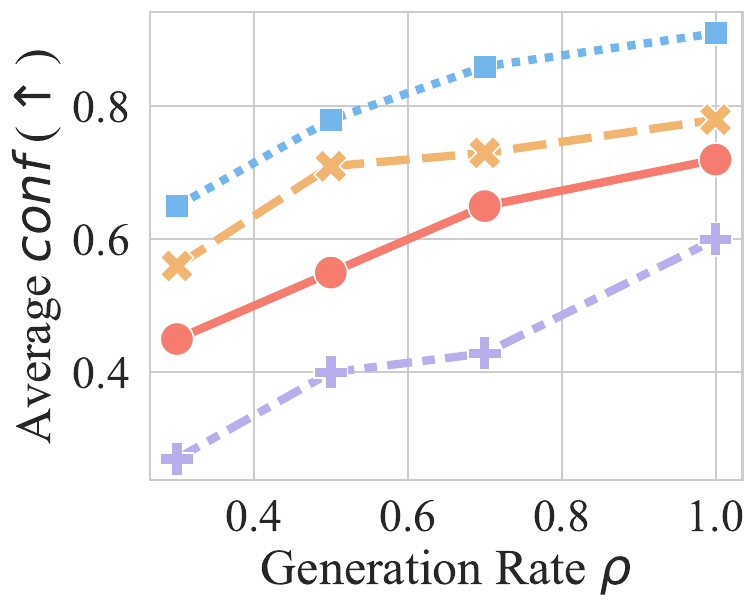}} \\
    \vspace{-3.5mm}   
    \subfigure[SD-v1-CC3M]{\includegraphics[width = 0.22\textwidth]{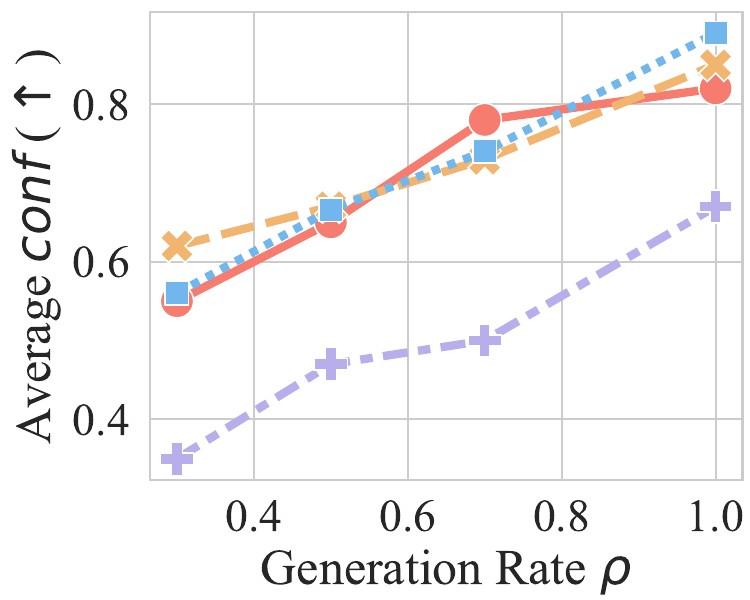}}
    \quad
    \subfigure[SD-v2-CC3M]{\includegraphics[width = 0.22\textwidth]{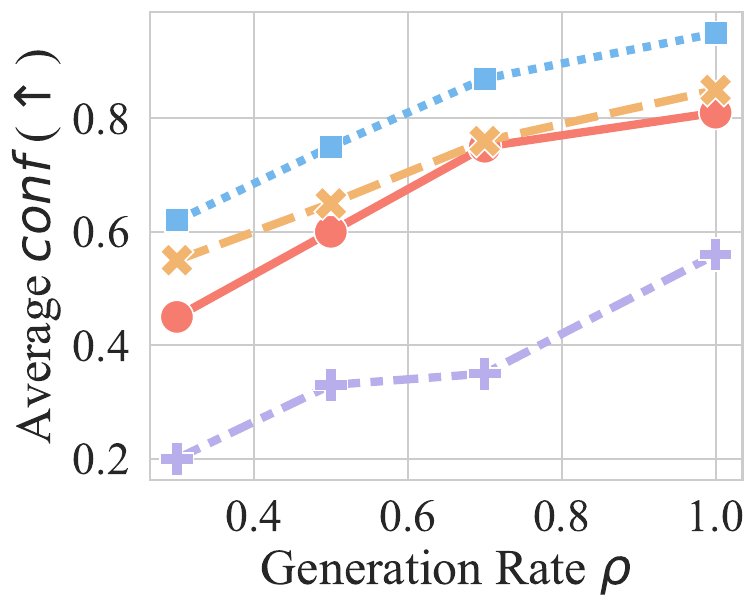}}\\
    \vspace{-3.5mm}
    \subfigure[SD-v1-CelebA]{\includegraphics[width = 0.22\textwidth]{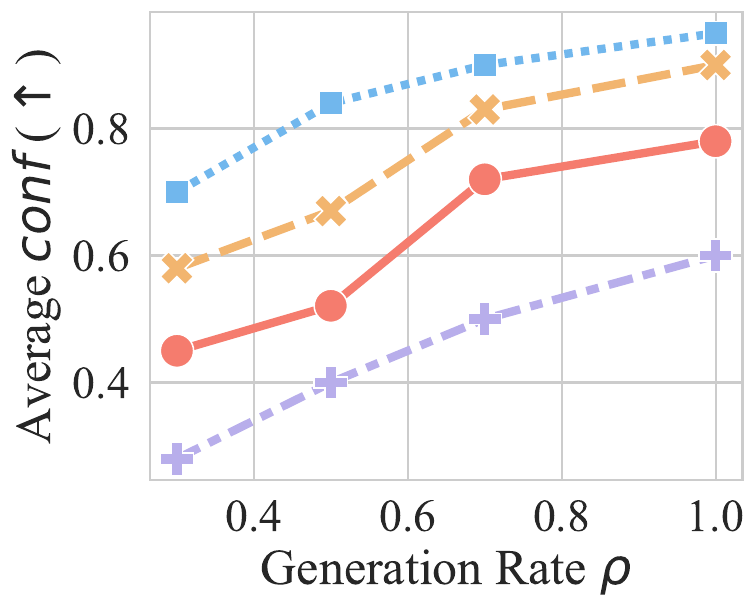}}
    \quad
    \subfigure[SD-v2-CelebA]{\includegraphics[width = 0.22\textwidth]{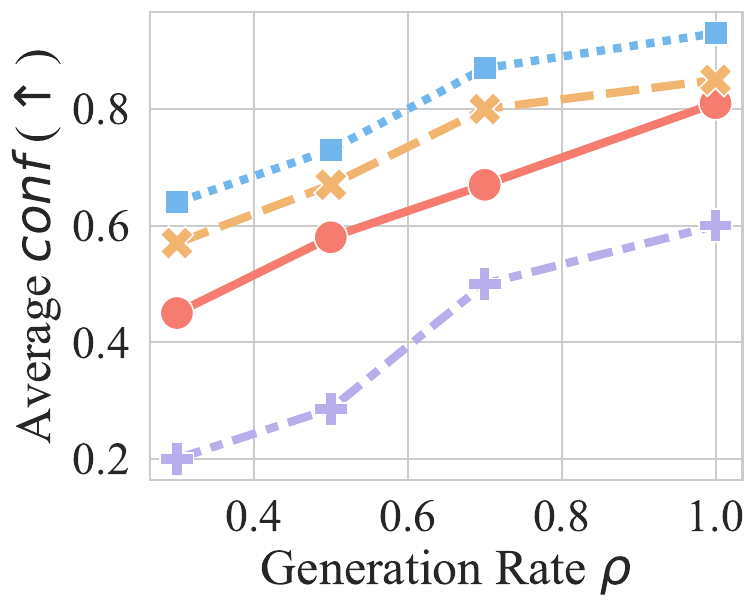}} 
    \caption{Effectiveness of instance-level attribution verification under different generation rate $\rho$.}
    \label{fig:rho}
\end{figure}

\begin{table}[t]
\centering
\caption{Performance of statistical-level attribution solution.}
\label{tab:shadow}
\resizebox{0.48\textwidth}{!}{
\begin{tabular}{c|c|ccc}
\hline
\multirow{2}{*}{Source Model} & \multirow{2}{*}{\begin{tabular}[c]{@{}c@{}}Pre-trained Model\\ of \\Infringing Model\end{tabular}} & \multicolumn{3}{c}{Metric} \\ \cline{3-5} 
 &  & \multicolumn{1}{c|}{Accuracy} & \multicolumn{1}{c|}{AUC} & 
 \begin{tabular}[c]{@{}c@{}}
 TPR@\\10\%FPR
 \end{tabular}
 \\ \hline
\multirow{2}{*}{SD-v1} & SD-v1 & \multicolumn{1}{c|}{0.90} & \multicolumn{1}{c|}{0.95} & 0.74 \\ \cline{2-5} 
 & SD-v2 & \multicolumn{1}{c|}{0.88} & \multicolumn{1}{c|}{0.91} & 0.78 \\ \hline
\multirow{2}{*}{SD-v2} & SD-v1 & \multicolumn{1}{c|}{0.91} & \multicolumn{1}{c|}{0.94} & 0.76 \\ \cline{2-5} 
 & SD-v2 & \multicolumn{1}{c|}{0.85} & \multicolumn{1}{c|}{0.93} & 0.75 \\ \hline
\multirow{2}{*}{SD-v1-CelebA} & SD-v1 & \multicolumn{1}{c|}{0.90} & \multicolumn{1}{c|}{0.92} & 0.81 \\ \cline{2-5} 
 & SD-v2 & \multicolumn{1}{c|}{0.94} & \multicolumn{1}{c|}{0.95} & 0.86 \\ \hline
\multirow{2}{*}{SD-v2-CelebA} & SD-v1 & \multicolumn{1}{c|}{0.93} & \multicolumn{1}{c|}{0.90} & 0.75 \\ \cline{2-5} 
 & SD-v2 & \multicolumn{1}{c|}{0.89} & \multicolumn{1}{c|}{0.84} & 0.74 \\ \hline
\multirow{2}{*}{SD-v1-CC3M} & SD-v1 & \multicolumn{1}{c|}{0.88} & \multicolumn{1}{c|}{0.81} & 0.78 \\ \cline{2-5} 
 & SD-v2 & \multicolumn{1}{c|}{0.91} & \multicolumn{1}{c|}{0.87} & 0.81 \\ \hline
\multirow{2}{*}{SD-v2-CC3M} & SD-v1 & \multicolumn{1}{c|}{0.86} & \multicolumn{1}{c|}{0.90} & 0.87 \\ \cline{2-5} 
 & SD-v2 & \multicolumn{1}{c|}{0.87} & \multicolumn{1}{c|}{0.88} & 0.83 \\ \hline
\end{tabular}}
\end{table}

\begin{table*}[!h]
\centering
\caption{Comparison of reconstruction distance distribution calculated with 30 key samples for the instance-level attribution solution. (Note that the innocent model does not have a corresponding source model.)}
\label{tab:delta0}
\setlength{\tabcolsep}{3.5mm}{\begin{tabular}{c|c|c|ccccc|c|c}
\hline
 & \multirow{2}{*}{\begin{tabular}[c]{@{}c@{}}Pretrain \\ Model\end{tabular}} & \multirow{2}{*}{\begin{tabular}[c]{@{}c@{}}Source\\Model\end{tabular}} & \multicolumn{5}{c|}{\begin{tabular}[c]{@{}c@{}}The Proportion of Samples \\ Within a Reconstruction Distance Range\end{tabular}} & \multirow{2}{*}{Average} & \multirow{2}{*}{Best} \\ \cline{4-8}
 &  &  & \multicolumn{1}{c|}{0-0.1} & \multicolumn{1}{c|}{0.1-0.15} & \multicolumn{1}{c|}{0.15-0.2} & \multicolumn{1}{c|}{0.25-0.3} & 0.3-0.4 &  &  \\ \hline
\begin{tabular}[c]{@{}c@{}}Innocent\\ Model\end{tabular} & SD-v1 & $-$ & \multicolumn{1}{c|}{0} & \multicolumn{1}{c|}{0.043} & \multicolumn{1}{c|}{0.739} & \multicolumn{1}{c|}{0.109} & 0.109 & 0.213 & 0.117 \\ \hline
\multirow{6}{*}{\begin{tabular}[c]{@{}c@{}}Infringing\\ Model\end{tabular}} & SD-v2 & SD-v1 & \multicolumn{1}{c|}{0.167} & \multicolumn{1}{c|}{0.472} & \multicolumn{1}{c|}{0.222} & \multicolumn{1}{c|}{0.139} & 0 & 0.151 & 0.033 \\
 & SD-v1 & SD-v2 & \multicolumn{1}{c|}{0.250} & \multicolumn{1}{c|}{0.458} & \multicolumn{1}{c|}{0.208} & \multicolumn{1}{c|}{0.084} & 0 & 0.148 & 0.067 \\
 & SD-v2 & SD-v1-CelebA & \multicolumn{1}{c|}{0.315} & \multicolumn{1}{c|}{0.424} & \multicolumn{1}{c|}{0.108} & \multicolumn{1}{c|}{0.103} & 0.050 & 0.113 & 0.019 \\
 & SD-v1 & SD-v2-CelebA & \multicolumn{1}{c|}{0.294} & \multicolumn{1}{c|}{0.324} & \multicolumn{1}{c|}{0.206} & \multicolumn{1}{c|}{0.090} & 0.086 & 0.121 & 0.026 \\
 & SD-v2 & SD-v1-CC3M & \multicolumn{1}{c|}{0.235} & \multicolumn{1}{c|}{0.457} & \multicolumn{1}{c|}{0.131} & \multicolumn{1}{c|}{0.175} & 0.002 & 0.133 & 0.042 \\
 & SD-v1 & SD-v2-CC3M & \multicolumn{1}{c|}{0.228} & \multicolumn{1}{c|}{0.495} & \multicolumn{1}{c|}{0.159} & \multicolumn{1}{c|}{0.118} & 0 & 0.125 & 0.035 \\ \hline
\end{tabular}}
\end{table*}



\subsection{Main Results}
\label{sec:results}

\noindent \textbf{Effectiveness of Instance-level Attribution.}
Given each source model, we built $30$ infringing models and calculated the $\texttt{conf}$ metric defined in \autoref{eq:conf} for each infringing model.
Here we set the key sample size as $N=30$.
To assess the reliability of our instance-level attribution solution, we report the average value of $\texttt{conf}$ among the $30$ infringing models under different generation rates $\rho$ in \autoref{fig:rho}.
The infringing models are fine-tuned with increasing proportions of generated images ($\rho=30\%, 50\%, 70\%, 100\%$ out of a total of $500$). 
The y-axis of \autoref{fig:rho} refers to the average $\texttt{conf}$ value. 
The higher the value, the more reliable our instance-level attribution solution is.

\underline{\textit{Main Result 1: }} 
Our solution surpasses Baseline 2, demonstrating a significant enhancement in attribution confidence by over 0.2 across diverse $\rho$ values.
Concurrently, our generation-based strategy for attribution attains a reliability equivalent to that of Baseline 1, with a minimal decrease in confidence not exceeding 0.1.


\underline{\textit{Main Result 2: }}
Our attribution method maintains its reliability even when the infringing model utilizes a small fraction of generated data for training. 
Our instance-level resolution, leveraging a generation-based strategy, exhibits a prediction confidence exceeding 0.6, even under a meager generation rate of 30\%. 
This performance illustrates a marked advantage, with a 50\% improvement over the baseline 2.

\noindent \textbf{Effectiveness of Statistical-level Attribution.}
To train the discriminator model in \autoref{sec:shadow}, we set $n=500, s=10, N=30$.
We evaluate the discriminator model and show the Accuracy, AUC, and TPR@10\%FPR metrics in \autoref{tab:shadow}.

\underline{\textit{Main Result 3:}} Results in \autoref{tab:shadow} show that our attribution achieves high accuracy and AUC performance, where the accuracy exceeds $85\%$, and AUC is higher than 0.8 for attributing infringing models to different source models. 
Accuracy and AUC are average-case metrics measuring how often an attribution method correctly predicts the infringement, while an attribution with a high FPR cannot be considered reliable.
Thus we use TPR@10\%FPR metric to evaluate the reliability of the statistical-level attribution.
The rightmost column of \autoref{tab:shadow} shows that TPR is over 0.7 at a low FPR of 10\%.
It means our attribution will not falsely assert an innocent model and is able to precisely distinguish the infringing models.




\subsection{Ablation Studies}
\label{ablation_study}

\begin{figure}[t]
    \centering
    \includegraphics[width=0.34\textwidth]{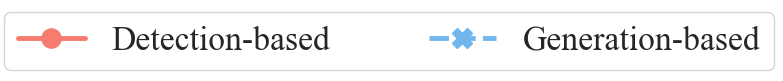}
    \vspace{-2.8mm}  
    \subfigure[SD-v1]{
    \centering\includegraphics[width = 0.22\textwidth]{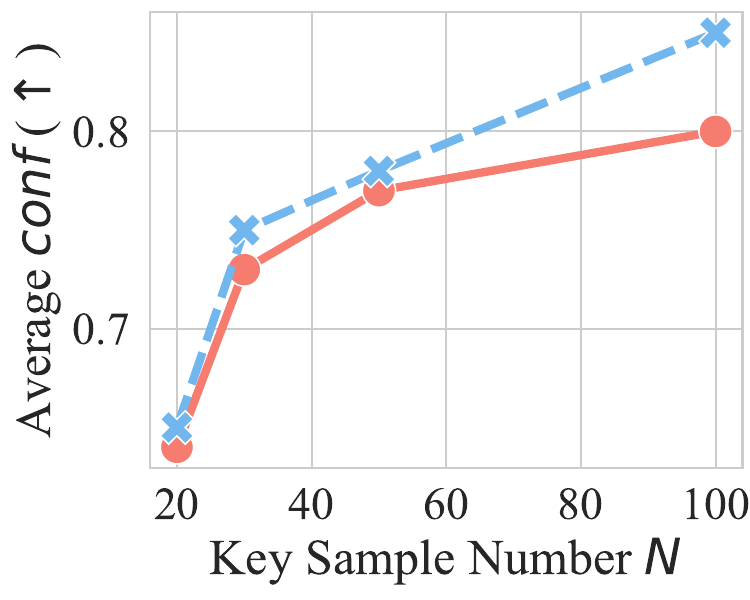}}
    \quad
    \subfigure[SD-v2]{\includegraphics[width = 0.22\textwidth]{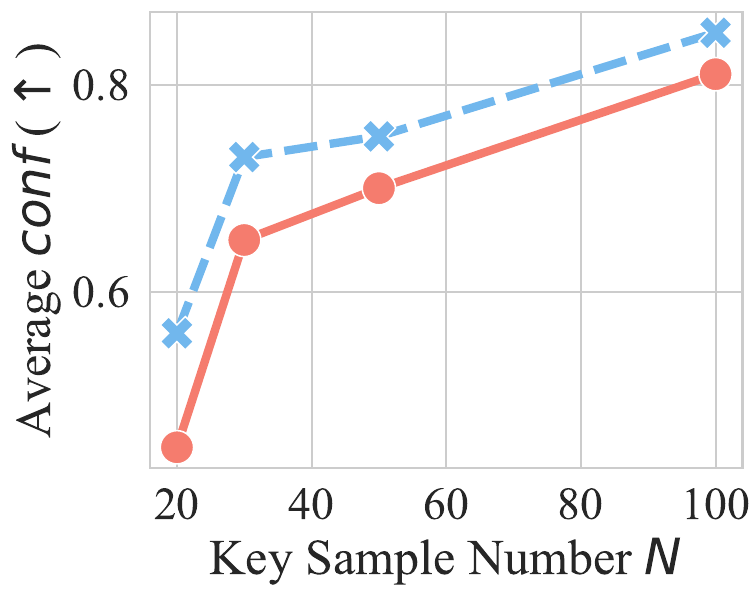}}  \\
    \subfigure[SD-v1-CC3M]{\includegraphics[width = 0.22\textwidth]{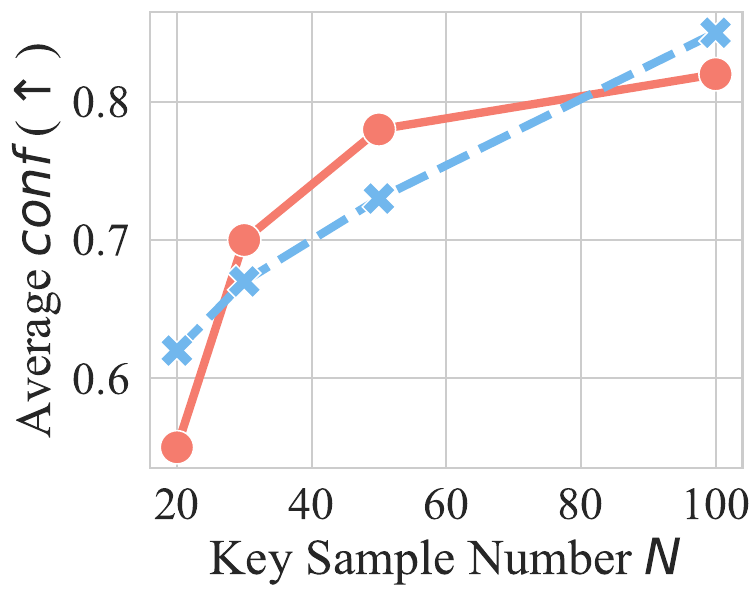}}
    \quad
    \subfigure[SD-v2-CC3M]{\includegraphics[width = 0.22\textwidth]{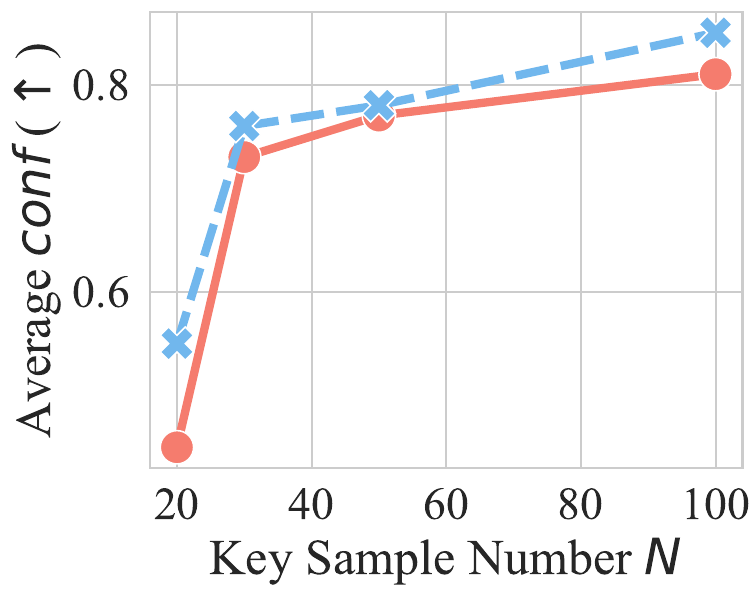}} \\
    \vspace{-3mm}
    \subfigure[SD-v1-CelebA]{\includegraphics[width = 0.22\textwidth]{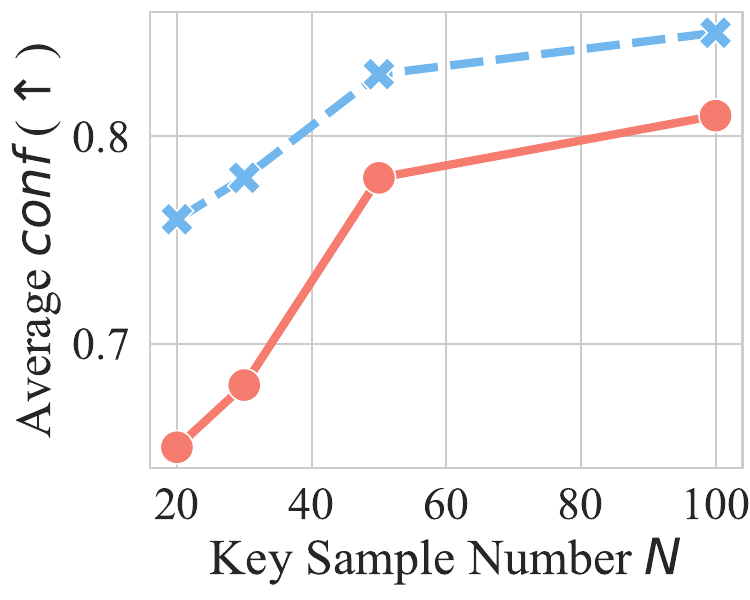}}
    \quad
    \subfigure[SD-v2-CelebA]{\includegraphics[width = 0.22\textwidth]{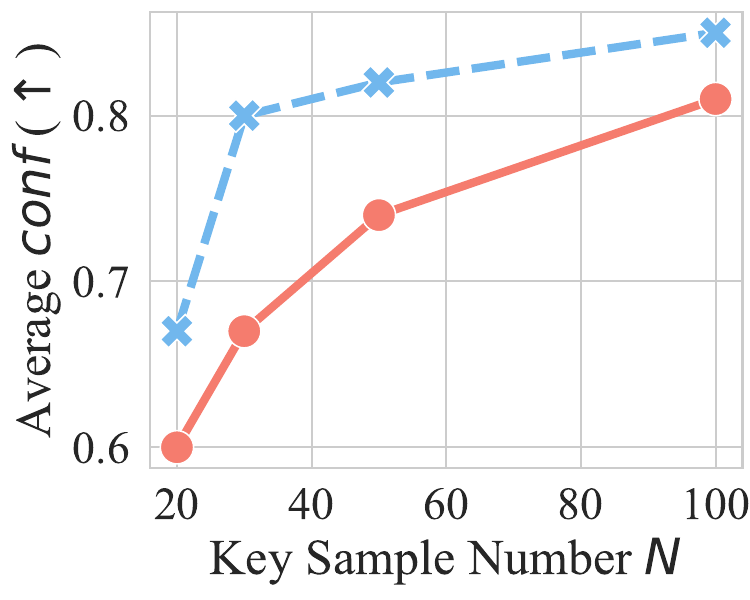}}
    \caption{Effectiveness of attribution verification under different key sample size $N$.}
    \vspace{-15pt}
    \label{fig:N}
\end{figure}

\noindent \textbf{Effect of hyper-parameter $\delta_0$.}
To determine an optimal value for $\delta_0$ for the instance-level attribution, we calculate the reconstruction distance values using $30$ key samples on an infringing model with $\rho=1$ and an innocent model with $\rho=0$.
The innocent model is finetuned on the pre-trained model of SD-v2.
\autoref{tab:delta0} compares the reconstruction distance distribution across the suspicious models based on different source models.
The columns 4-8 show the percentage of samples within a certain reconstruction distance range for each case, while the last 2 columns present the average and best reconstruction distance among all samples, respectively.
The more differences between the distributions of the innocent model and the infringing model, the easier to find a $\delta_0$ for attribution.

For the innocent model, the reconstruction distance of a large proportion of samples (as large as 73.9\%) falls within the range of [0.15,0.2), while only 4.3\% samples have reconstruction distance smaller than 0.15.
For the infringing model, there are about 20\% samples have reconstruction distance smaller than 0.1. In most cases (5 out of 6 infringing models), over a proportion of 40\% samples have the reconstruction distance within the range of [0.1,0.15). 
It indicates that $\delta_0=0.15$ is a significant boundary for distinguishing innocent models and infringing models regardless of the source models.
Hence, we set $\delta_0=0.15$ in our experiments.


\noindent \textbf{Effect of key sample size $N$.}
Following the settings in \autoref{tab:delta0}, we further study the impact of $N$ on the instance-level attribution, where $N$ ranges from 20 to 100 in \autoref{fig:N}.
The y-axis refers to the average value of $\texttt{conf}$ on the $N$ key samples through \autoref{eq:conf_f}, where $\texttt{conf}$ represents the attribution confidence to identify infringing models.
Each sub-figure in \autoref{fig:N} represents an infringing model with the corresponding source model specified in the sub-title.
The higher the confidence, the more reliable the attribution solution.
Theoretically, an increasing $N$ improves the verification reliability but requires more queries to the suspicious model.
Specifically, $N=100$ achieves the highest confidence, about 0.1 higher than that of $N=30$.
However, such a number of queries cause larger costs and worse stealthiness during the verification process.
$N=30$ has a similar performance as $N=50$, but it is dramatically better than $N=20$ with an advantage of about 0.1 when identifying the infringing models.
Thus in our above experiments, we set $N=30$ for efficiency.
In practice, the source model owner needs to make a trade-off between reliability and cost to set a suitable $N$.

\section{Conclusion}
\label{sec_concl}

This work tackles the crucial issue of training data attribution, investigating whether a suspicious model infringes on the intellectual property of a commercial model by using its generated data without authorization.
Our proposed attribution solution allows for the identification of the source model from which a suspicious model's training data originated.
The rationale of our method lies in leveraging the inherent memorization property of training datasets, which will be passed down through generated data and preserved within models trained on such data.
We devised algorithms to detect distinct samples that exhibit idiosyncratic behaviors in both source and suspicious models, exploiting these as inherent markers to trace the lineage of the suspicious model.
Conclusively, our research provides a robust solution for user term violation detection in the domain of text-to-image models by enabling reliable origin attribution without altering the source model's training or generation phase.


\bibliographystyle{ACM-Reference-Format}
\bibliography{sample-base}


\end{document}